\documentclass[12pt]{article}
\usepackage{amsmath}
\usepackage{amsfonts}
\usepackage{amssymb}
\usepackage{graphicx}
\usepackage{natbib}
\usepackage[colorlinks,citecolor=blue,urlcolor=blue,filecolor=blue,backref=page]{hyperref}
\usepackage{bm}
\usepackage{bbm}
\usepackage{xcolor}
\usepackage{subfig}
\usepackage{ulem}
\usepackage{enumitem}
\usepackage{authblk}
\usepackage{multirow}
\usepackage{tabularx}

\newcommand\bbE{\mathbb{E}} 
\newcommand\bbV{\mathbb{V}} 
\newcommand\bbP{\mathbb{P}} 
\newcommand{\bbR}{\mathbb{R}} 
\newcommand{\iidsim}{\overset{\text{iid}}{\sim}}
\newcommand{\indsim}{\overset{\text{ind}}{\sim}}

\newcommand{\bx}{{\bm x}}

\newcommand{\bz}{{\bm z}}

\newcommand{\bX}{{\bm X}}

\newcommand{\bw}{{\bm w}}

\newcommand{\bZ}{{\bm Z}}
\newcommand{\bW}{{\bm W}}

\newcommand{\bR}{{\bm R}}
\newcommand{\bI}{{\bm I}}
\newcommand{\bmu}{\bm \mu}
\newcommand{\bsigma}{\bm \sigma}
\newcommand{\bepsilon}{\bm \epsilon}
\newcommand{\bgamma}{{\bm \gamma}}
\newcommand{\btheta}{{\bm \theta}}

\newcommand{\bnu}{\bm \nu}

\newcommand{\sE}{\mathcal{E} }

\newcommand{\bzero}{{\bm 0}}
\newcommand{\bone}{{\bm 1}}
\newcommand\N{{\rm N}} 
\newcommand{\GP}{{\rm GP}} 
\newcommand{\tr}{{\rm tr}} 
\newcommand{\expit}{{\rm expit}} 

\newcommand\independent{\protect\mathpalette{\protect\independenT}{\perp}}
\def\independenT#1#2{\mathrel{\rlap{$#1#2$}\mkern2mu{#1#2}}}

\makeatletter
\newcommand*{\transpose}{%
	{\mathpalette\@transpose{}}%
}
\newcommand*{\@transpose}[2]{%
	\raisebox{\depth}{$\m@th#1\intercal$}%
}

\makeatother

\newcommand{\blind}{0}

\begin{document}

	\def\spacingset#1{\renewcommand{\baselinestretch}%
		{#1}\small\normalsize} \spacingset{1}

	
	\if0\blind
	{
		\title{\bf Non-stationary Gaussian process discriminant analysis with variable selection for high-dimensional functional data }
		\author{Weichang Yu\\
			Melbourne Centre for Data Science, University of Melbourne\\
			and \\
			Sara Wade\thanks{Royal Society of Edinburgh (RSE) Sabbatical Research Grant Holder; this work was supported by the RSE under Grant 69938.}\\
			School of Mathematics, University of Edinburgh\\
			
			and \\
			Howard D. Bondell\thanks{This work was supported by the Australian Research Council under grant FT190100374.}\\
			School of Mathematics and Statistics, University of Melbourne\\
			and \\
			
			Lamiae Azizi\\
			School of Mathematics and Statistics, University of Sydney\\
		}
		\maketitle
	} \fi
	
	\if1\blind
	{
		\bigskip
		\bigskip
		\bigskip
		\begin{center}
			{\LARGE\bf Title}
		\end{center}
		\medskip
	} \fi
	
	\bigskip
	\begin{abstract}
		High-dimensional classification and feature selection tasks are ubiquitous with the recent advancement in data acquisition technology. In several application areas such as biology, genomics and proteomics, the data are often functional in their nature and exhibit a degree of roughness and non-stationarity. These structures pose additional challenges to commonly used methods that rely mainly on a two-stage approach performing variable selection and classification separately. We propose in this work a novel Gaussian process discriminant analysis (GPDA) that combines these steps in a unified framework. Our model is a two-layer non-stationary Gaussian process coupled with an Ising prior to identify differentially-distributed locations. Scalable inference is achieved via developing a variational scheme that exploits advances in the use of sparse inverse covariance matrices. We demonstrate the performance of our methodology on simulated datasets and two proteomics datasets: breast cancer and SARS-CoV-2. Our approach distinguishes itself by offering explainability as well as uncertainty quantification in addition to low computational cost, which are crucial to increase trust and social acceptance of data-driven tools. 
		
	\end{abstract}

	\noindent%
	\vfill
	
	\newpage
	\spacingset{1.75} 
	\section{Introduction}
	
	Technological advances for collecting and measuring information in the biomedical field and beyond have led to an explosion in high-dimensional data. Such data can be used to identify patterns or markers and predict an outcome of interest. However, in fields such as genomics, proteomics, and chemometrics, the high-dimensional data is often functional and possesses complicated correlation structures. These complexities pose challenges to statistical and machine learning methods that are used for analyzing the data.
	
	As a motivating concrete example, we consider the context of predicting phenotypes and identifying biomarkers based on mass spectrometry (MS) data. MS technology measures the mixtures of proteins/peptides of tissues or fluids and produces an MS spectrum 
	\citep{cruz2008comparison}. The resulting experimental data consists of discretely observed functional spectra, with typically tens of thousands of observed locations and just a few hundred samples. Moreover,  data at neighboring locations tends to be highly correlated, with the strength of such correlations varying across the mass-to-charge ($m/z$) range. In addition, MS data tends to be noisy due to chemical noise, misalignment, calibration and other issues \citep{cruz2008comparison}. 
	
	To overcome such technical challenges,
	many approaches have been proposed which differ in how they deal with the complexity of the functional data. The most common approaches rely on a two-stage process, where the first stage reduces the dimension of the functional predictors (e.g. through functional principal component analysis), followed by the second stage which applies a classification technique on the reduced features  \citep[e.g.][]{hall2001functional, ferre2006multilayer}. More sophisticated techniques employ generalized functional linear regression to estimate the functional coefficients through basis expansion. Different choices of basis functions have been considered, including splines \citep{James2002,cardot2003spline}, wavelets \citep{brown2001bayesian}, and step functions \citep{Grollemund2018}. In addition,  regularization is typically employed to avoid overfitting, through prior distributions or penalty functions, including lasso \citep{zhao2012wavelet}, Bayesian variable selection priors \citep{zhu2010bayesian}, and random series priors \citep{li2018bayesian}.
	An overview is provided in \citet{reiss2017methods}. However, most of these methods do not perform variable selection to identify intervals or regions of the functional inputs which are relevant for predicting the class labels. Variable selection is important in this setting as retaining a large number of non-discriminative variables exacerbates the risk of overfitting. Moreover, variable selection allows 
	identification of disease markers and improves model interpretability. Relevant literature has demonstrated the importance of including a variable selection component in classification and prediction models with functional data, for example, fused lasso and its Bayesian analogues \citep{tibshirani2005sparsity, casella2010penalized}, Bayesian fused shrinkage  \citep{song2018bayesian}, and Bayesian sparse step functions \citep{Grollemund2018}. For multivariate functions of the input space, various extensions of penalizations or Bayesian shrinkage and spike-and-slab priors have been proposed to incorporate spatial smoothness in the functional coefficient and variable selection \citep[e.g][]{goldsmith2014,li2015spatial, kang2018scalar}.
	
	In this work, we focus on the discriminant analysis (DA) framework, which provides an alternative approach to generalized functional regression for classification tasks. In DA, the conditional distribution of the inputs $\bx$ given the class label $y$ is modelled, which is then flipped via Bayes theorem to obtain the classification rule for $y$ given $\bx$. The standard approach assumes that the class conditional distribution is a multivariate normal. A number of extensions have been developed  for high-dimensional functional data, including penalized linear DA \citep{hastie1995penalized} and functional linear DA \citep{james2001functional}. Notably, \cite{murphy2010variable} proposed a quadratic DA which incorporates variable selection through constraints on the covariance matrices, whereas \cite{ferraty2003} proposed a nonparametric DA based on kernel density estimation. In the Bayesian paradigm, \cite{stingo2011variable} developed a quadratic DA which includes latent binary indicators for variable selection with a Markov random field prior to incorporate known structures.    \cite{stingo2012bayesian} extended this by applying a wavelet transformation to account for smoothness in the functional inputs. \cite{GUTIERREZ201456} developed a robust Bayesian nonparametric (BNP) version of quadratic DA through a two-stage approach that first selects variables based on a fitted Gaussian process (GP) model for the functional data and individual quadratic DA across variables, and at the second stage, employs BNP mixture models for flexible discriminant analysis based on the selected variables. 
	
	In this article, we propose a novel and scalable Bayesian DA which performs variable selection and classification jointly, by combining recent developments in deep GPs \citep{dunlop2018deep} to flexibly model the functional inputs and incorporating Ising priors to identify differentially-distributed locations within a unified model framework. While powerful, GP models suffer from a well-known computational burden due to the need to store and invert large covariance matrices. Several approaches have been proposed in literature to enable scalability of GP models  \citep[e.g.][]{Kumar2009, Hensman2013, salimbeni2017doubly, Geoga2020}. In this work, we develop a scalable inference algorithm that  ameloriates the computational burden by utilizing the link between GPs and stochastic partial differential equations (SPDEs) to construct sparse precision matrices  \citep{Lindgren2011, grigorievskiy2017}  
	and combine various variational inference schemes.  
	Specifically, we focus on a two-level architecture of deep GPs with exponential covariance functions due to the attractive balance between flexibility and computational efficiency, and describe how to extend this framework to other settings. 
	We apply and demonstrate the utility of our model on proteomics-related mass spectrometry datasets, while also highlighting its relevance to other applications with functional data, including temporal gene expression \citep{leng2006classification}, nuclear magnetic resonance spectroscopy \citep{allen2013regularized}, chemometrics \citep{murphy2010variable}, agricultural production based temporal measurements \citep{Grollemund2018}, speech recognition \citep{hastie1995penalized}, and pregnancy loss based on hormonal indicators \citep{bigelow2009bayesian}, among others. Lastly, we emphasize that our proposed Bayesian model offers explainability as well as uncertainty quantification, which are desirable qualities for increasing trust and widespread acceptance of data-driven tools in biomedicine and other scientific fields.
	
	The paper is structured as follows. We present our proposed model in Section \ref{sec:model} and the developed posterior inference scheme in Section \ref{sec:inference}. In Section \ref{improveEfficiency}, we provide details of the implementation tricks required for  scalability and efficiency of the  algorithm.
	In Section \ref{sec:results}, we present numerical results to demonstrate the performance of our
	methodology in simulated and real datasets. Section \ref{sec:conc} summarizes and concludes with future directions.

	\section{Model}\label{sec:model}
	
	
	

	
	
	
	\begin{figure}[!t]
		\centering
		\subfloat[Controls]{\includegraphics[width=0.5\textwidth]{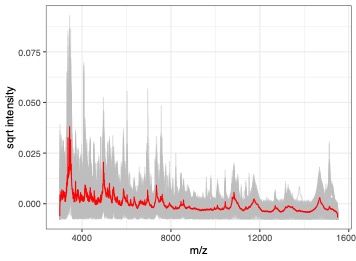}}
		\subfloat[SARS-CoV-2]{\includegraphics[width=0.5\textwidth]{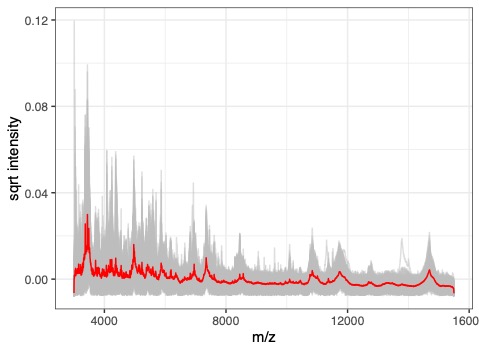}}\\
		\subfloat[Class-averaged spectra]{\includegraphics[width=0.65\textwidth]{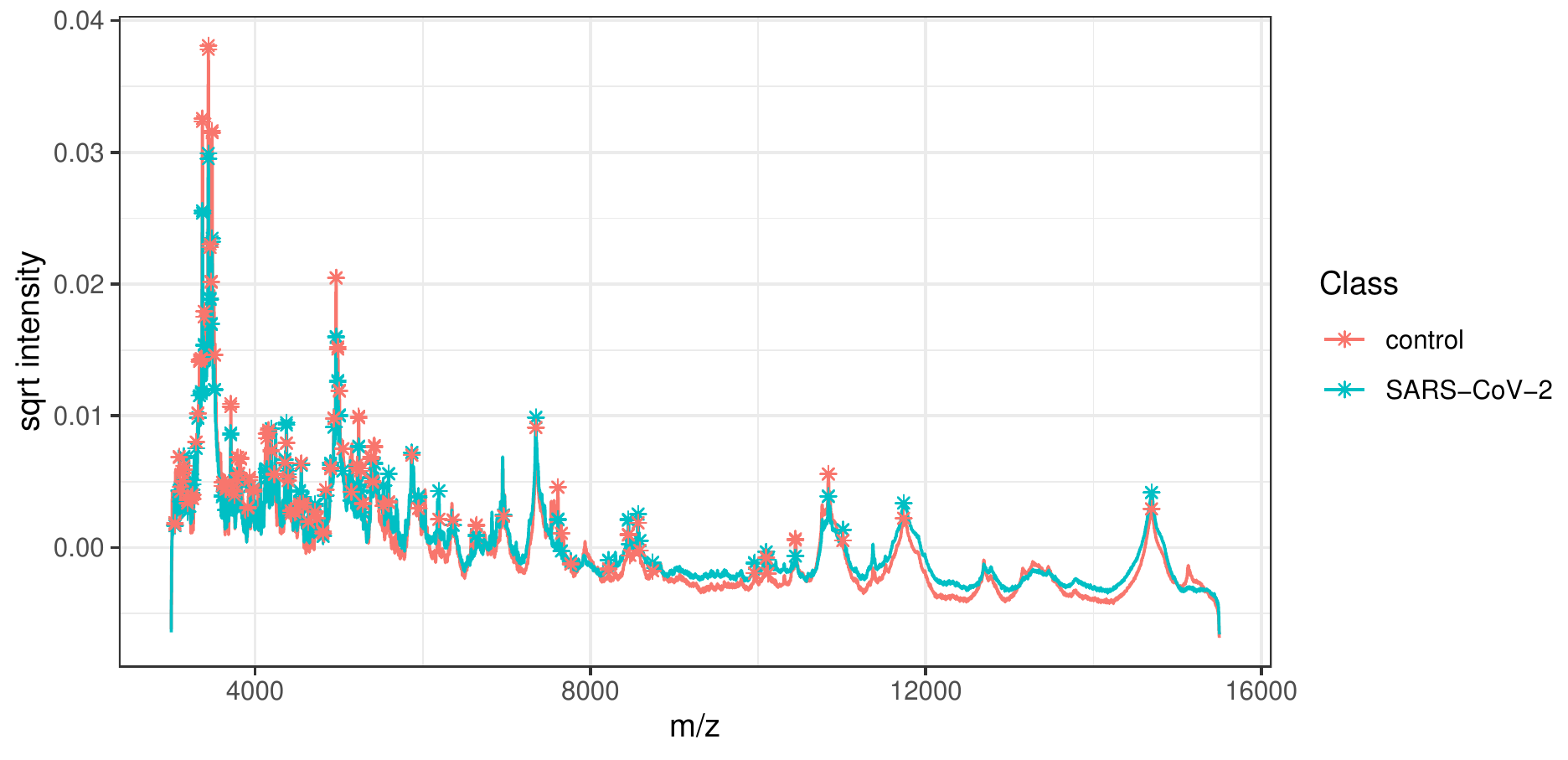}\label{fig:covid_avg}}
		\caption{Illustration of the SARS-CoV-2 data \citep{nachtigall2020detection}, with the square root intensities of the preprocessed spectra for (a) healthy controls and (b) SARS-CoV-2 positive patients (class-average given in red). The class averages are compared in (c) and the detected peaks, identified through the standard two-stage pipeline in \citet{nachtigall2020detection}, are marked with stars.}
		\label{fig:covid_spectra}
	\end{figure}
	Consider a set of $n$ 
	functional inputs 
	$\{ x_i (t)  \}_{i=1}^n$ defined on the domain $\mathcal{D} \subset \bbR$ and their corresponding class labels $\{ y_i \}_{i=1}^n$, where $x_i(t) \in \bbR$, $y_i \in \mathcal{Y}$, and $\mathcal{Y}$ is a set of class labels. 
	As a motivating example, Figure \ref{fig:covid_spectra} depicts  a set of processed MS spectra for healthy controls and SARS-CoV-2 positive patients \citep{nachtigall2020detection}, with the $x$-axis indicating the mass-to-charge ratio ($m/z$) and the $y$-axis indicating the intensity of the protein or peptide ions. The commonly used pipeline for analysing such data follows a two-stage approach. At the first stage, peak detection and extraction protocols are employed to yield a set of $m/z$ values and their intensity readings corresponding to the detected peaks (for SARS-CoV-2, the class-averaged spectra and detected peaks are compared in Figure \ref{fig:covid_avg}).  At the second stage, information from these peak regions are used to predict the outcome of interest and identify differentially-expressed proteins. 
	However, the initial steps of peak detection and extraction are critical for classification and may exclude important information from regions along the functional trajectory which 
	are useful for classification \citep{liu2009comparison}.
	
	
	A viable solution is to build a unified data analysis framework that utilizes the entire high-dimensional inputs and 
	incorporates variable selection within the model. In the context of DA, the classical vanilla model 
	is well-known to suffer from poor classification accuracy in high-dimensional datasets
	\citep{Bickel2004, Fan2008}. 
	The two-stage approach, which first selects variables according to specified criteria and then classifies with DA on the selected variables \citep[some examples in][]{Fan2008, DuarteSilva2011, Cui2015}, provides some improvements. However, information is lost when variable selection and classification are performed in separate stages. For example, two variables with adjusted p-values of 0.003 and 0.03 may be selected, but their difference in significance levels is not accounted for at the classification stage. Some proposed methods that circumvent this loss of information by incorporating variable selection directly within a DA model have been proposed have led to good classification performance in several examples \citep{Witten2011, Romanes2020}.
	
	In the same spirit, we propose a novel unified model that performs the variable selection and the classification steps simultaneously. 
	For brevity, we present our method in the context of a binary classification problem (although an extension to more than two classes is straightforward). Our proposed approach builds on a DA framework that employs GPs to flexibly model the entire functional trajectory: 
	\begin{equation}		\label{eqn::datadist}
	x_i \; \vert \; y_i=k, \mu_k, \sigma_k, \psi_i  \indsim  
	\GP(\mu_k, K_{\psi_i} + \sE_k ),
	\end{equation}
	where $y_i = k \in \lbrace 0,1 \rbrace$ refers to the class label; $\mu_k (t) \in \bbR$ is the group-specific mean function; $K_{\psi_i}$ is a covariance function (or kernel) with observation-specific parameters $\psi_i$; $\sE_k$ is a white noise kernel with class- and location-dependent variance $\sigma_k^2 (t) > 0$; and $\GP(\mu, K + \sE)$ denotes a Gaussian process with mean function $\mu$ and covariance function $K+\sE$. Here, $K$ and $\sE$ induce the marginal variance $\text{Var} \{ x_i(t) \} = K(t,t) + \sE(t)$ and the pairwise covariance $\text{Cov} ( x_i(t), x_i(t^\prime ) ) = K(t,t^\prime)$. Each stochastic process may be decomposed as
	$$
	x_i (t) = \mu_k (t) + z_i (t) + \epsilon_{k,i}(t),
	$$
	where 
	$z_i \mid \psi_i \sim \GP ( 0, K_{\psi_i})$ is an observation-specific latent process and $\epsilon_{k,i} (t) \in \bbR$ is a white-noise process with 
	variance $\sigma_k^2 (t) > 0$. Note the latent process $z_i$ accounts for the covariances between the values of the stochastic process at multiple locations, and $\epsilon_{k,i} (t)$ allows location-varying noisy errors, which are often present in functional data, such as the MS trajectories in Figure \ref{fig:covid_spectra}. We allow for variable selection in our proposed model by defining a binary signal process $\gamma (t) \in \{0,1\}$ such that
	\begin{equation*}
	\mu_k (t) = \gamma (t) \widetilde{\mu}_k (t) +  (1 - \gamma (t) ) \widetilde{\mu}_\emptyset (t) \;\; \text{and} \;\; \sigma_k^2 (t) = \gamma (t) \widetilde{\sigma}_k^2 (t) +  (1 - \gamma (t) ) \widetilde{\sigma}_\emptyset^2 (t),
	\end{equation*}
	where $\widetilde{\mu}_k$ and $\widetilde{\sigma}_k^2$ are the group-specific mean and noise variance processes at discriminative locations and $\widetilde{\mu}_\emptyset$ and $\widetilde{\sigma}_\emptyset^2$ are the common mean and noise variance processes at non-discriminative locations. In the context of MS data, $\gamma$ allows for detection of relevant $m/z$ values within the classification model. Thus, the entire aligned spectra are the inputs of the classification model, combining the steps of peak detection, feature extraction, and classification into a unified modeling framework. An example of a pair of $\mu_1$ and $\mu_0$ are depicted in the Supplementary Material. 
	The above parameterization allows us to directly infer the components $\widetilde{\mu}_k$ and $\widetilde{\mu}_\emptyset$ for the mean functions and $\widetilde{\sigma}_k^2$ and $\widetilde{\sigma}_\emptyset^2$ for the noise variance functions, as well as the signal process $\gamma$, which determines the regions where the mean and variance differ between groups. 

	

\begin{figure}[!t]
		\centering
		\subfloat[Common log length-scale $R(t)$]{\includegraphics[width=0.5\textwidth]{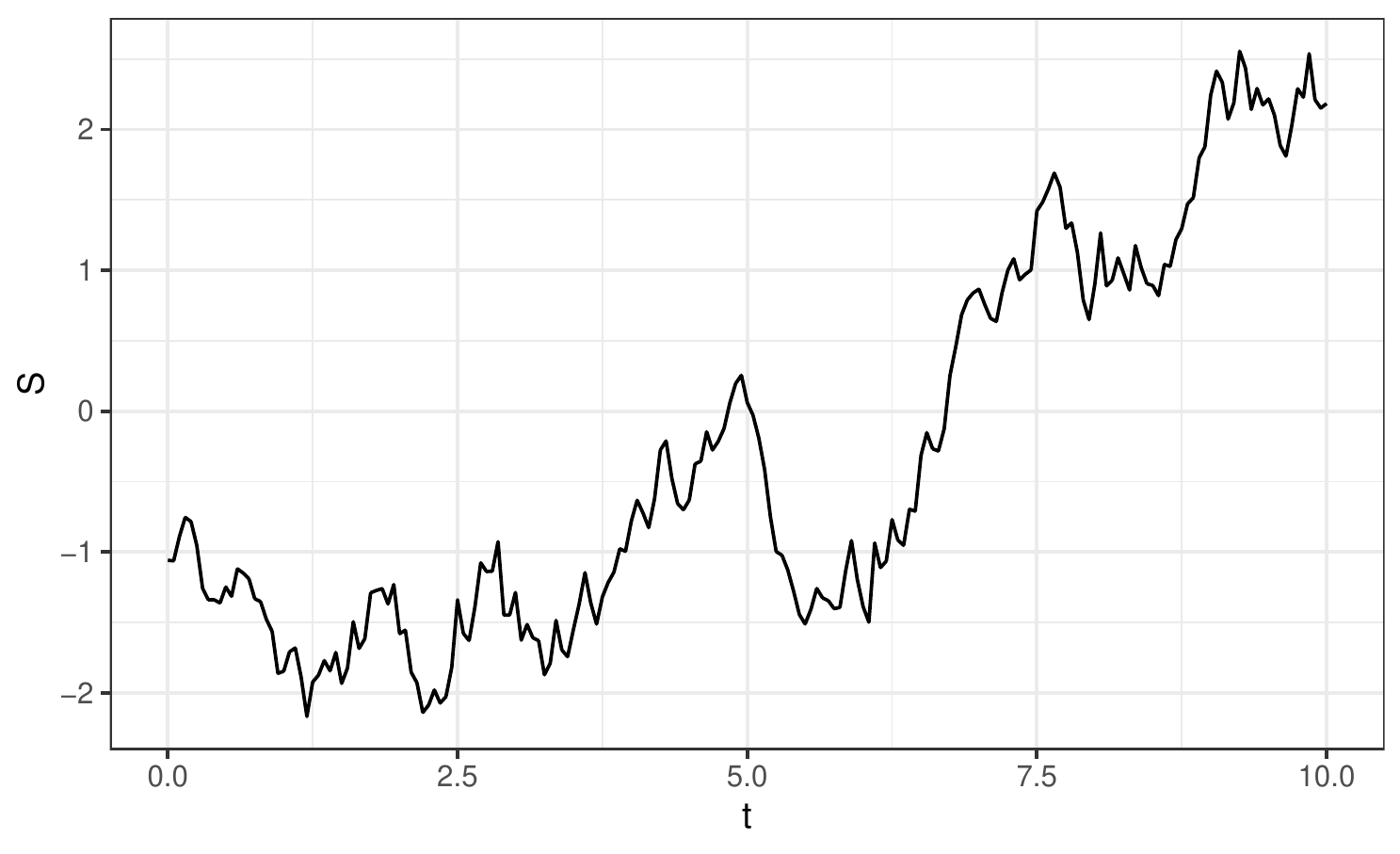}} 
		\subfloat[Non-stationary process $z_i(t)$]{\includegraphics[width=0.5\textwidth]{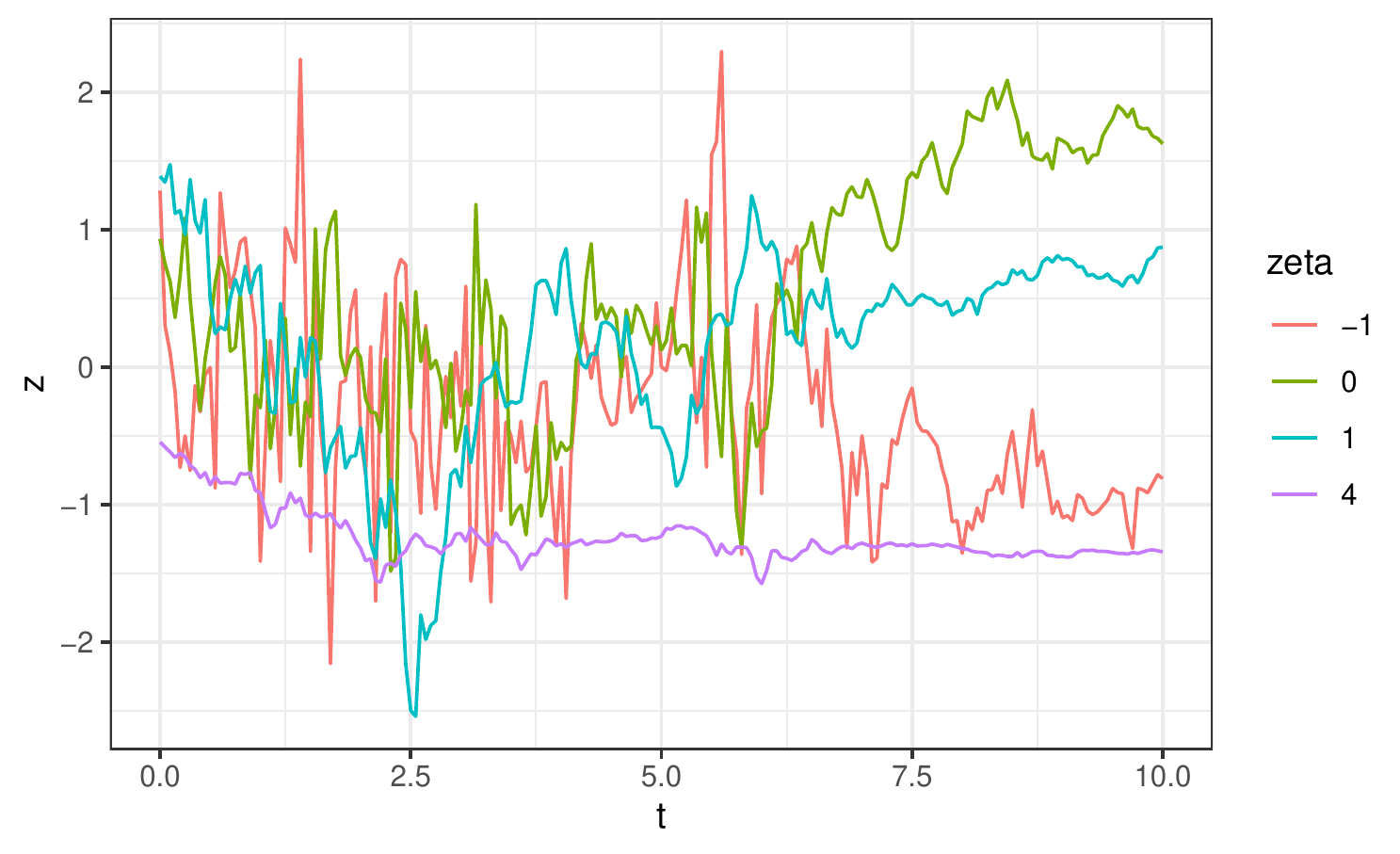}} 
		\caption{Illustration of the two-level non-stationary GP. Larger/smaller values of the common log-length scale process $R(t)$ in (a) result in  flatter/more wiggly behavior for all observed latent $z_i(t)$. In addition, the log-length scale process for each observation is perturbed by $\zeta_i$, allowing for some $z_i(t)$ (e.g. with $\zeta_i = 4$) to be flatter than others across the entire domain. }
		\label{fig:ns_illus}
	\end{figure}

	\paragraph{Two-level non-stationary Gaussian processes.} 
	In several examples such as in our motivating example of MS data, the observed functions are often unevenly rough which is indicative of a non-stationary covariance structure. This behavior is evident in Figure \ref{fig:covid_spectra}, where the spectra are flatter in some regions and change more rapidly in others. To account for this behavior, we assign a non-stationary covariance kernel \citep{Paciorek2003} for $K_{\psi_i}$. 
	Specifically, the kernel parameter, $\psi_i = (\tau, \nu_i)$, consists of the magnitude $\tau>0$ and a location-varying log length-scale process $\nu_i$, i.e., $K_{\psi_i} = K_{NS;\tau, \nu_i}$, and hence we may write
	\begin{align*}
	z_i | \tau, \nu_i \sim \GP(0, K_{NS;\tau, \nu_i}).
	\end{align*}
	At the second level, we place Gaussian process priors on the log length-scale processes with
	$$\nu_{i} (t) = R(t) + \zeta_i, \quad \text{ and } \quad R \sim \GP(\mu_\nu, K_{S;\tau_2, \lambda}),$$
	where $K_{S;\tau_2, \lambda}$ is a stationary covariance kernel with marginal scale $\tau_2$ and length scale $\lambda$. Here, each observation-specific log length-scale process has been decomposed into a common component $R(t)$ to account for the location-varying covariance structure common across all observed functions, and an observation-specific perturbation $\zeta_i \in \bbR$ to allow for between-spectra variation in smoothness across the entire domain. This behavior is illustrated in Figure \ref{fig:ns_illus}. This decomposition is flexible yet also reduces the computational cost of having to infer $n$ observation-specific location-varying log length-scales (details are provided in the Supplementary Material).
	
	Due to the observed roughness in our motivating dataset (refer to Figure \ref{fig:covid_spectra}), we focus on the exponential covariance function, a member of the Mat\'ern family with smoothness parameter $\nu=1/2$. In one-dimension, this is also known as the Ornstein-Uhlenbeck process, and it is the continuous-time counterpart of the first-order autoregressive model AR(1). Motivated by the link between GPs and SPDEs \citep{Lindgren2011, monterrubio2018}, we employ an SDE representation of the nonstationary processes: 
	\begin{align}
	dz_i &= -\frac{1}{\exp(\nu_i)} z_i dt + \sqrt{\frac{2 \tau}{\exp(\nu_i)}} d \omega_{1},
	\label{eq:SDElev1}\\
	dR &= -\frac{1}{\lambda} R dt + \sqrt{\frac{2 \tau_{2}}{\lambda}} d \omega_{2}, \label{eq:SDElev2}
	\end{align}
	where $ \nu_{i} (t) = R(t) + \zeta_i$ and $\omega_{1}$ and $\omega_{2}$  are Wiener processes.  
	While appropriate for rough MS data, 
	the SDE representation can be extended to other choices of covariance functions, such as Mat\'ern processes with smoother realisations \citep{grigorievskiy2017}. 
	The resulting two-level GP model 
	provides the flexibility needed to capture non-stationarities, but this could be further extended with a non-stationary covariance kernel also for the log length-scale process by using deeper architectures \citep{dunlop2018deep,zhao2020deep}. 
	

	\subsection{Finite-time discretization}\label{sec:finite}
	In practice, the functional realizations $\bx_i = (x_i(t_1), \ldots, x_i(t_T))$ are discretely-observed at locations $t_1 < \ldots < t_T$\footnote{For ease of notation, we assume the locations $t_1 < \ldots < t_T$ are common across all observations, but the discretized model could be extended accordingly.}. Following our proposed model for the process $x_i$ in \eqref{eqn::datadist}, the likelihood of the discretely-observed vector $\bx_i$ is a multivariate normal density:
	$$
	\bx_i \; \vert \; y_i, \bmu_{y_i}, \bnu_i, \tau, \bsigma_{y_i}^2  \sim  \left\{ \begin{array}{ll}
	\N(\bmu_1, Q_{NS;\tau, \bnu_i} ^{-1} + D_{\epsilon1} ), & \mbox{ if $y_i = 1$;} \\ [1ex]
	\N(\bmu_0, Q_{NS;\tau, \bnu_i} ^{-1} + D_{\epsilon0} ), & \mbox{ if $y_i = 0$},
	\end{array} \right.
	$$
	where $\bmu_k = ( \mu_k(t_1), \ldots, \mu_k (t_T ))^\top$; $\bsigma_{k}^2  = ( \sigma_k^2 (t_1), \ldots, \sigma_k^2 (t_T)  )^\top$ ; $ \bnu_i = \zeta_i + \bR $ with $\bR = (R(t_1),\ldots, R(t_T))^\top$; $D_{\epsilon k}  = \text{dg} ( \sigma_k^2 (t_1), \ldots, \sigma_k^2 (t_T)  )$; dg forms a diagonal matrix from elements of a vector, and $Q_{NS;\tau, \bnu_i}$ is a precision matrix parameterized $\tau$ and $\bnu_i$. Again, the discretely-observed $\bx_i$ can be decomposed as:
	$$\bx_i = \bmu_k + \bz_i + \bepsilon_{k,i}, \quad \text{ for } y_i = k,$$
	where $\bz_i = (z_i(t_1), \ldots, z_i(t_T) )^\top $ is the observation-specific latent vector with precision matrix $Q_{NS;\tau, \bnu_i}$;  $\bepsilon_{k,i}$ is the white noise vector with covariance matrix $ D_{\epsilon k}$;  
	$$ \bmu_k = \bgamma \odot \widetilde{\bmu}_k + (\bone-\bgamma) \odot \widetilde{\bmu}_\emptyset, \quad \text{ and } \quad \bsigma_k = \bgamma \odot \widetilde{\bsigma}_k + (\bone-\bgamma) \odot \widetilde{\bsigma}_\emptyset,$$
	where $\odot$ denotes element-wise product. At the second level, the common log length-scale vector $\bR$ has a multivariate normal distribution,
	$$ \bR  \, \vert \, \lambda, \tau_2 \sim \N( \bzero, Q_{S; \lambda, \tau_2}^{-1} ). $$  
	
	To overcome the computational bottleneck of GPs, we construct sparse precision matrices $Q_{NS;\tau, \bnu_i}$ and $Q_{S; \lambda, \tau_2}$ by discretizing the SDE representations in \eqref{eq:SDElev1} and \eqref{eq:SDElev2}, respectively, on the grid $t_1 < \ldots < t_T$ using the Euler-Maruyama scheme:
	\begin{align}
	z_i(t_j) &= z_i (t_{j-1}) -\frac{1}{\exp(\nu_i(t_{j-1}))} z_i(t_{j-1}) \delta_j + \sqrt{\frac{2 \tau}{\exp(\nu_i(t_{j-1}))}} w_{1,j}, \label{eq:em1}\\
	R(t_{j}) &= R(t_{j-1}) -\frac{1}{\lambda} R(t_{j-1}) \delta_j + \sqrt{\frac{2 \tau_{2}}{\lambda}}  w_{2,j}, \label{eq:em2}
	\end{align}
	where $\nu_i(t_{j}) = \zeta_i + R(t_{j}) $; $w_{1,j} \sim \text{N}(0, \delta_j) \independent  w_{2,j} \sim \text{N}(0, \delta_j) $; and $\delta_j = t_j - t_{j-1}$. 
	For notational simplicity, we assume the locations are equally-spaced, i.e., $t_j - t_{j-1} = \delta$. Through \eqref{eq:em1} and \eqref{eq:em2}, the precision matrices of $\bz_i$ 
	and $\bR$ have a tridiagonal structure (specific form in the Supplementary Material), which alleviates the complexity and enhances the  efficiency of our posterior inference algorithm. A detailed discussion is provided in Section \ref{improveEfficiency}.

	\subsection{Choice of priors} 
	To reflect our prior belief that the underlying variable selection process $\gamma$ is smooth, we assign a linear chain Ising prior \citep{li2010} to account for smoothness. Following this choice of prior, the conditional distribution of $\gamma(t)$ given its corresponding set of neighbors with locations  in $\mathcal{N}_t \subset \lbrace t_1, \ldots, t_T \rbrace$ is 
	\begin{align*}
	\bbP ( \gamma (t) = 1 \, \vert \, \{ \gamma (t^\prime) \}_{t^\prime \in \mathcal{N}_t} ) = \expit \left  \{ -\alpha +  \sum_{t^\prime \in \mathcal{N}_t} \beta(t, t^\prime) \gamma (t^\prime)   \right \},
	\end{align*}
	where $\expit(x) = \{1+ \exp(-x) \}^{-1}$, $\alpha \in \bbR$, and $\beta(t,t^\prime) > 0$. Here, a larger value of $\alpha$ corresponds to more sparsity in $\gamma$, whereas $\beta$ controls the correlation and smoothness between the values of $\gamma$ at neighboring locations. 
	In our context, we define $\mathcal{N}_t= \lbrace t-1, t+1 \rbrace$ and $\beta(t,t') =\beta$.  
	
\begin{figure}[!h]
		\centering
		{\includegraphics[width=0.8\textwidth]{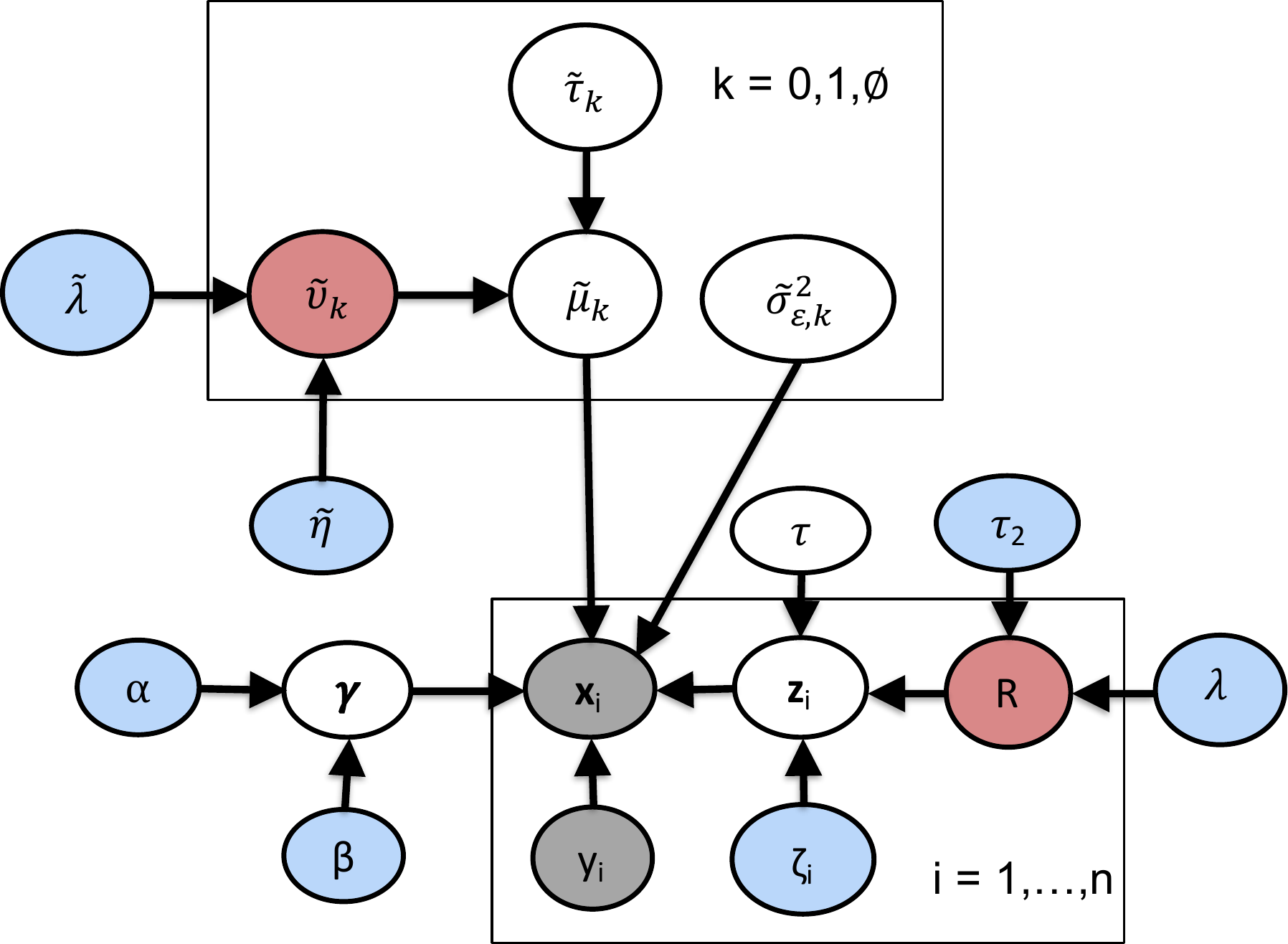}}
		\caption{DAG representation of the proposed model. The colour fills denote the inference approach adopted: blue for MAP; red for SVB; white for CAVI; and gray denotes observed data. More details of the various inference approaches is provided in Section \ref{sec:inference}. }
		\label{fig:DAG}
	\end{figure}

	Motivated by the non-stationary behavior also evident in the class-averaged functions (see Figure \ref{fig:covid_spectra}), we also assign  hierarchical GP priors for the mean functions
	\begin{align*}
	\widetilde{\mu}_k (t) \, \vert \,  \widetilde{\tau}_k, \widetilde{\nu}_k \, &\sim \, \GP(0, K_{NS;  \widetilde{\tau}_k, \widetilde{\nu}_k}), \\
	\widetilde{\nu}_k (t) \, \vert \, \widetilde{\eta}, \widetilde{\lambda}  \; &\sim \; \GP (\mu_{\widetilde{\nu}} , K_{S; \widetilde{\eta}, \widetilde{\lambda} } ),
	\end{align*}
	with hyperpriors $\widetilde{\tau}_k \sim \text{InvGa}( A_{\widetilde{\tau}}, B_{\widetilde{\tau}} )$ for $k = 0, 1, \emptyset$;  
	$\widetilde{\eta} \sim \text{InvGa}(A_{\widetilde{\eta}},B_{\widetilde{\eta}})$; and  $\widetilde{\lambda} \sim \text{LogN}( \mu_{\widetilde{\lambda}}, \sigma_{\widetilde{\lambda}}^2)$. Here, the hyperparameters of the length scale $\widetilde{\lambda}$ are specified such that the resultant prior is weakly informative to mitigate potential non-identifiability issues  \citep[][and further details in the Supplementary Material]{Betancourt2017robust}. 
	For the noise variance, we assign an independent prior at each location
	$$
	\widetilde{\sigma}_k^2 (t) \, \sim \, \text{InvGa} (A_\epsilon, B_\epsilon), \text{ for } k =0,1, \;\; \text{and} \;\; \widetilde{\sigma}_\emptyset^2 (t) \, \sim \, \text{InvGa} (A_\epsilon, B_\epsilon).
	$$
	For the magnitude $\tau$ of the $z_i$ and the magnitude $\tau_2$ of $\bR$, we assume $\tau \sim \text{InvGa}(A_\tau, B_\tau)$ and $\tau_2 \sim \text{InvGa}(A_{\tau_2}, B_{\tau_2})$, respectively. The observation-specific perturbations are assigned the priors $\zeta_i \iidsim \text{N}(0, \sigma_\zeta^2)$, where $\sigma_\zeta^2=1$. For the length scale $\lambda$, we assign the weakly informative hyperprior $\lambda \sim \text{LogN}( \mu_{\lambda}, \sigma_{\lambda}^2)$. 
	A directed acyclic graph (DAG) that summarizes the parameters in our proposed model is provided in Figure \ref{fig:DAG}.

	\section{Inference and prediction}\label{sec:inference}
	While Markov chain Monte Carlo is a popular technique for computing posterior distributions in Bayesian modelling,
	an alternative approximate Bayesian inference method known as \textit{variational Bayes} has gained popularity in the literature. Variational Bayes has
	been shown to be a fast posterior computation method that yields reasonably accurate approximations in several problems. 
	Consider fitting a model parameterized by $\btheta$ to the observed data $\mathcal{D}$. 
	In variational Bayes, the actual posterior $p(\btheta \, \vert \, \mathcal{D})$ is approximated by a density $q(\btheta)$ from a family of distributions $\mathcal{F}$ that maximizes the evidence lower bound (ELBO)
	\begin{equation}
	\label{eqn::ELBO}
	\mathbb{E}_{q(\btheta) } \left [ \log \left \{ \frac{p(\btheta,  \mathcal{D})  }{q(\btheta)}  \right\} \right ].
	\end{equation}

	A common choice for $\mathcal{F}$ is the mean-field family on the partition $\{ \btheta_1, \ldots, \btheta_L \}$ of $\btheta$:
	$$
	q(\btheta) = \prod_{l=1}^L q_l(\btheta_l),
	$$
	where $L \le \text{dim} (\btheta)$. Without any further parametric assumptions, it has been shown \citep{Ormerod2010} that the optimal choice for each product component $q_l$ is
	\begin{equation}
	\label{eqn::CAVIupdate}
	q_l (\btheta_l) \propto \exp \left [ \bbE_{- \btheta_l} \log \{  p(\btheta,  \mathcal{D}) \}  \right ],
	\end{equation}
	where the above expectation is taken with respect to $\prod_{l^\prime \neq l} q_{l^\prime} (\btheta_{l^\prime})$. This choice of product component is known as \textit{coordinate ascent variational inference} (CAVI). Note that in some cases, the RHS in \eqref{eqn::CAVIupdate} is intractable. In such cases, we adopt alternative methods such as stochastic variational Bayes (SVB) or maximum a posteriori (MAP)  to circumvent the intractability.

	\subsection{Posterior inference}\label{sec:vb}
	Let $\btheta$ denote the vector of all model parameters (excluding the hyperparameters $\zeta_i$, $\lambda$, $\tau_2$, $\widetilde{\lambda}$, $\widetilde{\eta}$, $\alpha$, and $\beta$). 
	We specify the mean-field family for the approximate posterior:
	\begin{align*}
	q(\btheta) &= q(\tau) q(\bR) \prod_{i=1}^n \{ q(\bz_i)  \}   \times \prod_{k \in \{ \emptyset, 0, 1 \} } \left  \{ q(\widetilde{\bmu}_k ) q( \widetilde{\bnu}_k) q( \widetilde{\tau}_k) \prod_{j=1}^T q(\widetilde{\sigma}_{kj}^{2})  \right  \}   \times \prod_{j=1}^T \{ q(\gamma_j) \}.
	\end{align*}
	In the following, we provide details on the functional form of each product component. Note that the product components for the location-varying log length-scales, i.e. $\bR$ and $\bnu_i$, are computed using SVB updates, while CAVI updates are employed for  all other parameters in $\btheta$. The remaining  hyperparameters $\zeta_i$, $\lambda$, $\tau_2$, $\widetilde{\lambda}$, $\widetilde{\eta}$, $\alpha$, and $\beta$ are optimized using  MAP estimation,  with the marginal likelihood approximated by the ELBO \eqref{eqn::ELBO}. In Figure \ref{fig:DAG}, parameters in white, red, and blue fill are updated with CAVI, SVB and MAP respectivly, whereas gray fill denotes observed quantities. For the rest of this section, we use the notation $\bbE$ to denote an expectation with respect to the variational posterior and a subscript $j$ to denote the value of the functional at location $t_j$, e.g. $\widetilde{\mu}_k (t_j) = \mu_{kj}$ and $\gamma_j = \gamma(t_j)$. Furthermore, as proof of concept, we present our approximate posteriors in the case whereby each stochastic process $x_i$ is observed at the same set of equally-spaced locations $t_1, \ldots, t_T$, with $t_{j+1} - t_j = \delta$. 
	
	\paragraph{Mean function parameters.} The product component for $\widetilde{\bmu}_k$ is:
	\begin{equation*}
	q(\widetilde{\bmu}_k) = \text{N} (m_{\widetilde{\bmu}_k}, Q_{\widetilde{\bmu}_k}^{-1} ), 
	\end{equation*}
	where, for $k =0,1$ and $k = \emptyset$, respectively, we have
	\begin{align*}
	&m_{\widetilde{\bmu}_k} = Q_{\widetilde{\bmu}_k}^{-1} \bbE ( \widetilde{D}_{\epsilon k}^{-1} ) \bW  \left ( \bX^{(k)} - m_{\bZ}^{(k)} \right )^\top \bone_{n_k} \; \text{and} \; Q_{\widetilde{\bmu}_k} = n_k \mathbf{W} \bbE ( \widetilde{D}_{\epsilon k}^{-1} ) + \bbE Q_{NS; \widetilde{\tau}_{k}, \widetilde{\boldsymbol{\nu}}_k }, \\
	&m_{ \widetilde{\bmu}_\emptyset} = Q_{\widetilde{\bmu}_\emptyset}^{-1} \bbE ( \widetilde{D}_{\epsilon \emptyset}^{-1} )  (\bI - \bW)  \left ( \bX - m_\bZ \right )^\top \bone_n\; \text{and} \; Q_{\widetilde{\bmu}_\emptyset} = n (\mathbf{I} - \mathbf{W} ) \bbE ( \widetilde{D}_{\epsilon \emptyset}^{-1} ) + \bbE Q_{NS; \widetilde{\tau}_{\emptyset}, \widetilde{\boldsymbol{\nu}}_\emptyset };
	\end{align*}
	with 
	$n_k$ denoting the number of training observations in group $k = 0, 1$ and $\bone_n$ denotes a column-vector of size $n$ with all entries equal to one. Here, $\bw = \bbE (\bgamma)$, $\bW = \text{dg} (  \bw)$, and $m_{\bZ} = \bbE ( \bZ)$ with $\bZ = [\bz_1, \ldots, \bz_n]^\top$, $\bX = [\bx_1, \ldots, \bx_n]^\top$, and $\bX^{(k)}$ and $m_{\bZ}^{(k)}$ denoting the rows of $\bX$ and $m_{\bZ}$, respectively, corresponding to training observations from class $k$.
	The product component for the magnitude  $\widetilde{\tau}_{k}$ is:	
	\begin{equation*}
	q(\widetilde{\tau}_{k}) = \text{InvGa} (a_{\widetilde{\tau}_k}, b_{\widetilde{\tau}_k} ),
	\end{equation*}
	where $a_{\widetilde{\tau}_k} = A_{\widetilde{\tau}} + T/2$ and $ b_{\widetilde{\tau}_k} = B_{\widetilde{\tau}} + \text{\tr}[ \bbE( \widetilde{\bmu}_k \widetilde{\bmu}_k^\top) \bbE C_{NS; \widetilde{\boldsymbol{\nu}}_k }]/2$, and $\tr$ denotes the trace operator. Here, $C_{NS;  \widetilde{\boldsymbol{\nu}}_k } = \widetilde{\tau}_{k}Q_{NS; \widetilde{\tau}_{k}, \widetilde{\boldsymbol{\nu}}_k }$ represents the non-stationary precision matrix with unit marginal scale 
	(computational details in the Supplementary Material). 
	For the location-varying log length-scale $\widetilde{\bnu}_k$,  non-conjugacy between the priors for $\widetilde{\bnu}_k$ and  $\widetilde{\bmu}_k$ leads to an intractable CAVI update. To circumvent this intractability, we adopt SVB and specify a Gaussian  form for $q(\widetilde{\bnu}_k)$, i.e.,
	$$
	q(\widetilde{\bnu}_k) = \text{N}(m_{\widetilde{\bnu}_k}, (\Omega_{\widetilde{\bnu}_k}\Omega_{\widetilde{\bnu}_k}^{\top})^{-1} ),
	$$
	where $m_{\widetilde{\bnu}_k}$ and $\Omega_{\widetilde{\bnu}_k}$ are chosen to maximize the evidence lower bound:
	\begin{align*}
	&\text{ELBO} (m_{\widetilde{\bnu}_k},\Omega_{\widetilde{\bnu}_k}) =\bbE \left [ \log p(\widetilde{\bmu}_k \, \vert \, \widetilde{\bnu}_k) + \log p(\widetilde{\bnu}_k) - \log q(\widetilde{\bnu}_k) \right ] \\
	&=\tfrac{1}{2} \bone^\top \bbE( \widetilde{\bnu}_k) - \tfrac{1}{2} m_{1/\tau} \text{\tr}[ \bbE( \widetilde{\bmu}_k \widetilde{\bmu}_k^\top) \bbE C_{NS; \widetilde{\boldsymbol{\nu}}_k }]\\ 
	&- \tfrac{1}{2} \bbE \left\{ (\widetilde{\bnu}_k - \mu_{\widetilde{\nu}} )^\top 
	Q_{S;\widetilde{\eta},\widetilde{\lambda}}
	(\widetilde{\bnu}_k - \mu_{\widetilde{\nu}} ) \right \} - \bbE \log q(\widetilde{\bnu}_k; m_{\widetilde{\bnu}_k}, \Omega_{\widetilde{\bnu}_k}),
	\end{align*}
	and $m_{1/\widetilde{\tau}_k} = \bbE(1/ \widetilde{\tau}_k )$.
	To reduce the computational complexity, 
	we specify a sparse Cholesky decomposition for the variational precision matrix. In particular, $\Omega_{\widetilde{\bnu}_k}$ is a 1-banded lower triangular matrix, leading to a tridiagonal precision matrix. 
	Note that, in addition to the computational savings, this form is further motivated as the approximate posterior precision matrix maintains a similar structure to the prior precision matrix.  
	Details about the steps for this SVB update are provided in the Supplementary Material.
	The magnitude $\widetilde{\eta}$ and length-scale $\widetilde{\lambda}$ of the log length-scale processes are updated as MAP estimates, i.e., the maximizer of the objective
	\begin{equation*}
	\text{OBJ} (\widetilde{\eta},\widetilde{\lambda}) = \sum_{k \in \{0,1, \emptyset\}} \bbE \log \phi( \widetilde{\bnu}_k ; \mu_{\widetilde{\bnu}}, Q_{S; \widetilde{\eta}, \widetilde{\lambda} }^{-1} ) + \log p( \widetilde{\eta} ) + \log p(\widetilde{\lambda}).
	\end{equation*}
	
	\paragraph{Noise variance parameters.} The product component for $\widetilde{\sigma}_k^2$ is:
	\begin{equation*}
	q(\widetilde{\sigma}_{kj}^2) = \text{InvGa} (r_{kj}, s_{kj}). 
	\end{equation*}
	Here, for $k = 0,1$ and $k = \emptyset$, respectively, we have
	\begin{align*}
	r_{kj}  = A_{\epsilon} + n_kw_j/2, \quad &\text{ and } \quad
	s_{kj} = B_\epsilon + \tfrac{w_j}{2} \sum_{i:y_i=k} \bbE (x_{ij} - \widetilde{\mu}_{kj} - z_{ij})^2; \\
	r_{\emptyset j}  = A_{\epsilon} + n(1-w_j)/2, \quad &\text{ and } \quad
	s_{\emptyset j} = B_{\epsilon} + \tfrac{1-w_j}{2} \sum_{i=1}^n \bbE (x_{ij} - \widetilde{\mu}_{\emptyset j} - z_{ij})^2, 
	\end{align*}
	where,
	$\bbE (x_{ij} - \widetilde{\mu}_{kj} - z_{ij})^2 = (x_{ij} - m_{\widetilde{\mu}_{kj}} - m_{z_{ij}} )^2 + (Q_{\widetilde{\bmu}_k}^{-1})_{jj}  + (Q_{\bz_i}^{-1})_{jj}$, $\bbE ( \widetilde{\mu}_{kj} ) = m_{\widetilde{\mu}_{kj}}$,  $\bbV( \widetilde{\bmu}_{k} ) =  Q_{\widetilde{\bmu}_k}^{-1} $, $\bbV ( \bz_i )  = Q_{\bz_i}^{-1}$, and the subscript $jj$ denotes the $(j,j)$-th entry of a matrix. 
	
	\paragraph{Latent process.} The product component for $\bz_i$:
	\begin{equation*}
	q(\bz_i) = \text{N} (m_{\bz_i}, Q_{\bz_i}^{-1}),
	\end{equation*}
	where
	\begin{align}
	&Q_{\bz_i} = \bW \bbE ( \widetilde{D}_{\epsilon y_i}^{-1}) + ( \bI - \bW ) \bbE (\widetilde{D}_{\epsilon \emptyset}^{-1}) +  \bbE (Q_{NS;\tau, \boldsymbol{\nu}_i})  , \label{eq:zvar}\\
	&m_{\bz_i} = Q^{-1}_{\bz_i} \left \{ \bW \bbE ( \widetilde{D}_{\epsilon y_i}^{-1}) (\bx_i - m_{ \widetilde{\bmu}_{y_i}}) + (\bI - \bW) \bbE (\widetilde{D}_{\epsilon \emptyset}^{-1}) (\bx_i - m_{ \widetilde{\bmu}_{\emptyset}}) \right \}. \label{eq:zmean}
	\end{align}
	\paragraph{Covariance parameters.} For the magnitude $\tau$ of the latent process, 
	$ q(\tau) = \text{InvGa} (r_\tau, s_\tau)
	$ 
	where, $r_\tau = A_{\tau} + nT/2$ and
	\begin{align*}
	s_\tau = b_\tau + \tfrac{1}{2} \left[ \sum_{i=1}^n \text{\tr}\left(\bbE[\bz_{i}\bz_{i}^\top] \bbE[ C_{NS; \bnu_i}]\right)   
	\right],
	\end{align*}
	with $C_{NS; \bnu_i} = \tau Q_{NS; \tau,  \bnu_i}$ denoting the precision matrix with unit precision. For the common log length-scale vector $\bR$, 
	we adopt a SVB update  by specifying its approximate density:
	\begin{equation*}
	q(\bR) = \N( m_{\bR}, (\Omega_{\bR}  \Omega_{\bR}^{ \top})^{-1}  ),
	\end{equation*}
	where the variational parameters $m_{\bR}$ and $\Omega_{\bR}$ are chosen to maximize: 
	\begin{align*}
	\text{ELBO}(m_\bR, \Omega_{\bR}) = \bbE \Big [ \sum_{i=1}^n \log \phi (\bz_{i}; \bzero, Q_{NS;\tau, \boldsymbol{\nu}_i}^{-1} ) + \log \phi(\bR; \bzero,  Q_{S;\tau_2, \lambda}^{-1} )-\log q(\bR; m_\bR, \Omega_{\bR})  \Big ].
	\end{align*}
	The details of the SVB update are similar to the update for $\widetilde{\bnu}_k$ and thus are omitted. The observation-specific perturbations $\zeta_i$ are updated as the maximizers of the objective:
	\begin{align*}
	\text{OBJ}(\zeta_i) = \bbE \Big [ \log \phi (\bz_{i}; \bzero,  Q_{NS;\tau, \boldsymbol{\nu}_i}^{-1} )   \Big ] + \log p( \zeta_i ).
	\end{align*}
	The magnitude $\tau_2$ and length-scale $\lambda$ of the common location-varying log length-scale $\bR$ are updated as MAP estimates, i.e.,  maximizers of the objective
	\begin{equation*}
	\text{OBJ} (\tau_2,\lambda) =  \bbE \log \phi( \bR ; 0, Q_{S; \tau_2, \lambda}^{-1} ) +  \log p( \tau_2 ) + \log p(\lambda).
	\end{equation*}
	
	\paragraph{Feature selection parameter.} For the binary indicator $\gamma_j$ at each location,  
	\begin{align*}
	w_j &= q(\gamma_j = 1) = \expit \left [ - \frac{u_{j}}{2} - \tfrac{1}{2} \bone^\top \mathbf{g}_j - \alpha + \beta (w_{j+1} + w_{j-1})    \right ],
	\end{align*}
	where $u_j = \tfrac{n_1}{2} \bbE \log (\widetilde{\sigma}_{1j}^{2}) + \tfrac{n_0}{2} \bbE \log \widetilde{\sigma}_{0j}^{2}) - \tfrac{n}{2} \bbE \log (\widetilde{\sigma}_{\emptyset j}^{2})$, $\mathbf{g}_j = (g_{1j}, \ldots, g_{n j})^\top$, and
	\begin{align*}
	g_{ij} &= \bbE(\widetilde{\sigma}_{y_i j}^{2}) \left \{ (x_{ij} - m_{ \widetilde{\mu}_{y_i} j} - m_{z_{i} j})^2 + (Q_{ \widetilde{\bmu}_{y_i}}^{-1})_{jj} + (Q_{\bz_i}^{-1})_{jj}  \right \} \\
	&
	-  \bbE(\widetilde{\sigma}_{\emptyset j}^{2}) \left \{ (x_{ij} - m_{ \widetilde{\mu}_{\emptyset} j} - m_{z_{i}j})^2 + (Q_{\widetilde{\bmu}_\emptyset}^{-1})_{jj} + (Q_{\bz_i}^{-1})_{jj}  \right \}.
	\end{align*}
	The hyperparameters $\alpha$ and $\beta$ controlling sparsity and correlation of the Ising prior are updated as the maximizer of the objective:
	\begin{align*}
	&\text{OBJ}(\alpha,\beta)= \bbE \left [ \log p(\bgamma \, \vert \, \alpha, \beta) + \log p(\alpha) + \log p(\beta) \right ], 
	\end{align*}
	where in the case of the linear chain Ising prior, a closed form expression is available for the partition function (see \cite{salinas2001} and details in the Supplementary Material). 
	
	\subsection{Classification}
	Upon convergence of the variational parameters in the posterior inference phase, we proceed to derive a classification rule for a new process $x_{n+1} (t)$ that follows the distribution as described in equation \eqref{eqn::datadist}.
	This requires the predictive distribution $p(y_{n+1}, \bz_{n+1} \, \vert \, \mathcal{D}, \bx_{n+1})$, where $\mathcal{D}$ denotes all observed data. To simplify computations, we make the following mean field approximation for the predictive distribution of $y_{n+1}$ and $\bz_{n+1}$:
	$$
	p(\bz_{n+1}, y_{n+1} \, \vert \, \mathcal{D}, \bx_{n+1}) \approx q(\bz_{n+1}) q(y_{n+1}).
	$$
	Following equation (\ref{eqn::CAVIupdate}), we adopt a CAVI update for $\bz_{n+1}$, i.e.,
	\begin{align*}
	q(\bz_{n+1}) = \N(m_{\bz_{n+1}}, \Sigma_{\bz_{n+1}} ),
	\end{align*}
	where $m_{\bz_{n+1}}$ and $\Sigma_{\bz_{n+1}}$ are similar in form to the variational updates for $\bz_{i}$ in \eqref{eq:zvar}-\eqref{eq:zmean} but also take into account the unknown class label $y_{n+1}$: 
	\begin{align*}
	&Q_{\bz_{n+1}} =   \bW \left \{ \xi_1 \bbE \widetilde{D}_{\epsilon 1}^{ -1} + (1-\xi_1) \bbE \widetilde{D}_{\epsilon 0}^{ -1}  \right \} + (\bI - \bW) \bbE \widetilde{D}_{\epsilon \emptyset}^{ -1} + \bbE (Q_{NS;\tau, \boldsymbol{\nu}_{n+1}}), \\ &m_{\bz_{n+1}} = Q_{\bz_{n+1}}^{-1} \bW \left \{ \xi_1 \bbE \widetilde{D}_{\epsilon 1}^{-1} (\bx_{n+1} - m_{\widetilde{\bmu}_1} )  + (1-\xi_1) \bbE \widetilde{D}_{\epsilon 0}^{-1} (\bx_{n+i} -  m_{\widetilde{\bmu}_0})  \right \} \\ 
	&\hspace{1.5cm} + Q_{\bz_{n+1}}^{-1} (\bI - \bW) \bbE \widetilde{D}_{\epsilon \emptyset}^{ -1} (\bx_{n+i} -  m_{\widetilde{\bmu}_\emptyset}),
	\end{align*}
	where $\xi_1 = q(y_{n+1}=1)$. Note that $Q_{\bz_{n+1}}$ depends on the estimate of the observation-specific log length-scale $v_{n+1} (t) = R(t) + \zeta_{n+1}$. Since the approximate posterior for $R$ has been computed in the posterior inference phase, we only need an estimate for the perturbation $\zeta_{n+1}$. This may be computed via MAP estimation as the maximizer of:
	\begin{align*}
	\text{OBJ}(\zeta_{n+1}) = \bbE \Big [ \log \phi (\bz_{n+1}; \bzero, Q_{NS;\tau, \boldsymbol{\nu}_{n+1}}^{-1} )   \Big ] + \log p( \zeta_{n+1} ).
	\end{align*}

	For the group label $y_{n+1}$, we adopt a CAVI update:
	\begin{align*}
	&\xi_1 = q(y_{n+1} = 1) \\
	&= \expit \Bigg [  - \tfrac{1}{2} \bw^\top \bbE  \left \{ \log (\widetilde{\bsigma}_{1}^{2} ) -   \log (\widetilde{\bsigma}_{0}^{2}) \right \} - \tfrac{1}{2} \text{QDA} (\bx_{n+1}) - \tfrac{1}{2} \tr \left \{ \bW \bbE(\widetilde{D}_{\epsilon 1}^{-1}) (  Q_{\widetilde{\bmu}_1}^{-1} + Q_{\bz_{n+1}}^{-1} )  \right \}  \\
	&+ \tfrac{1}{2} \tr \left \{ \bW \bbE(\widetilde{D}_{\epsilon 0}^{-1}) (  Q_{\widetilde{\bmu}_0}^{-1} + Q_{\bz_{n+1}}^{-1} )  \right \} + \log \left ( \frac{n_1}{n_0} \right ) \Bigg ],
	\end{align*}
	where $\log (\widetilde{\bsigma}_{k}^{2} )$ denotes the element-wise log of the vector $\widetilde{\bsigma}_{k}^{2}$ and
	\begin{align*}
	\text{QDA} (\bx_{n+1}) &= (\bx_{n+1} - m_{\widetilde{\bmu}_1} - m_{\bz_{n+1}})^\top \bW \bbE( \widetilde{D}_{\epsilon 1}^{ -1} ) (\bx_{n+1} - m_{\widetilde{\bmu}_1} - m_{\bz_{n+1}}) \\
	&\hspace{0.5cm}  - (\bx_{n+1} - m_{\widetilde{\bmu}_0} - m_{\bz_{n+1}})^\top \bW \bbE( \widetilde{D}_{\epsilon 0}^{ -1} ) (\bx_{n+1} - m_{\widetilde{\bmu}_0} - m_{\bz_{n+1}}).
	\end{align*}	
	The posterior inference and classification phases of the algorithm are terminated when the change in values of the variational parameters and MAP estimates are sufficiently small. Pseudocodes for the posterior inference and classification algorithms are provided in Algorithms S1 and S2 in the Supplementary Material.
	
	
	
	\section{Improving Scalability and Efficiency}
	\label{improveEfficiency}
	A na{\"i}ve implementation of the inference algorithm described in the previous section would be impractical as it involves computationally costly operations with large matrices. For example, 
	the variational parameters for the latent processes $\bz_i$ 
	require solving  systems of $T$ linear equations for the evaluation of $m_{\bz_i}$ and computing the diagonals of $Q_{\bz_i}^{-1}$, with a total cost of $O(nT^3)$ for both operations. Moreover, if a full variational precision matrix is specified for the location-varying length-scales, the SVB updates may be slow to converge as the variational posterior would have $O(T^2)$ variational parameters . In the motivating SARS-CoV-2 example with $T=25,001$, these steps would clearly be infeasible.
	
	In this section, we briefly describe the computational shortcuts that we have adopted to reduce the computational complexity of our entire inference algorithm from $O(nT^3)$ to $O(nT)$, with further details in the Supplementary Material. 
	In particular, a careful inspection of all steps in the variational algorithm reveals  that we can avoid the costly $O(nT^3)$ operations and only require two types of operations which admit computationally efficient implementations: (1) solve a system of linear equations of the form $\mathbf{Q} \mathbf{a} = \mathbf{b}$, where $\mathbf{Q}$ is a $T \times T$ tridiagonal matrix and $\mathbf{a}, \mathbf{b} \in \mathbb{R}^T$; (2) computing the main and first off-diagonal entries of the inverse tridiagonal precision matrix. This can be attributed to both the tridiagonal structure of the precision matrices obtained through the specified kernel and SDE discretization, detailed in Section \ref{sec:finite}, and the sparse Cholesky decomposition of the variational precision matrix for the location-varying log length-scales, specified in Section \ref{sec:vb}. Additionally, the banded form of the Cholesky decomposition reduces the number of variational parameters in the SVB updates to $O(T)$.

	To compute efficiently the main diagonal and first off-diagonal entries, we adopt the sparse inverse subset algorithm \citep{Takahashi1973, durrande2019}. The algorithm begins by computing the inverse of the 1-banded Cholesky decomposition of the precision matrix, which also turns out to be a lower triangular 1-banded matrix, and then utilizes a recursive algorithm for computing the required entries. For solving the required system of linear equations, we utilize Thomas' algorithm \citep{Higham2002} that exploits efficiently the tridiagonal structure of $\mathbf{Q}$ and results in a low computational complexity $O(T)$. 
	Lastly, we note that for applications with smoother functional realizations, Mat\'ern kernels with a larger smoothness parameter can be used. This also leads to sparse banded precision matrices but with a higher bandwidth $b$ and increased computational complexity of $O(b^2T)$. 
	
	\section{Numerical results}\label{sec:results}
	In this section, we study the performance of our proposed Gaussian process DA (\texttt{GPDA}) in four simulation settings and two publicly available proteomics datasets. We compare with seven other methods - variational nonparametric DA \citep[\texttt{VNPDA,}][]{Yu2020CSDA}, penalized linear DA with fused lasso penalty \citep[\texttt{penLDA-FL},][]{Witten2011}, random forest, sparse linear DA \citep[\texttt{SparseLDA},][]{Clemmensen2011}, variational linear DA \citep[\texttt{VLDA},][]{Yu2020SC}, and both the $L_2$-regularized and $L_1$-regularized support vector machine (\texttt{SVM}) with linear kernels \citep{Cortes1995, Fan2008Liblinear}. For the proteomics datasets, we also compare with the traditional two-stage algorithm which involves peak detection and  followed by classification using linear DA and quadratic DA. These classifiers are selected as competing methods as their implementations are publicly available, have reasonably low computation time for high-dimensional datasets. Moreover, they perform variable selection except the $L_2$-regularized support vector machine.

	
	\subsection{Simulated datasets}
	In all simulations, we set $T=5000$. Each simulation setting is repeated 50 times. At each repetition, we draw a training dataset of size $n_{train}=100$ and a testing dataset of size $n_{test}=500$ from the distribution:
	$$
	\bx_i \; \vert \; y_i, \btheta^\star \sim  \left\{ \begin{array}{ll}
	\N_T ( \bmu_1^\star , \boldsymbol{\Sigma}_{i1}^\star ), & \mbox{ if $y_i = 1$; and} \\ [1ex]
	\N_T ( \bmu_0^\star , \boldsymbol{\Sigma}_{i0}^\star  ), & \mbox{ if $y_i = 0$},
	\end{array} \right.
	$$
	where the superscript $\star$ denotes the simulation setting. 
	The class labels are generated from: $y_i \sim \text{Bernoulli} (0.5)$. 
	Full details on the simulation settings for the mean function and noise variances are provided in the Supplementary Material.
	\paragraph{Description.} In {\textbf{Simulation 1}}, we study the performance of the methods when a large proportion (40\%) of the locations have weak predictive power, whereas the rest of the locations do not have any predictive power. The \texttt{GPDA} model is correctly specified, i.e., the covariance function of the $i$-th observation is
	$
	\boldsymbol{\Sigma}_{ik}^\star = D_{\epsilon, k}^\star +  Q_{NS; \tau^\star, \bnu_i^\star}^{-1}, \;\; \text{and} \;\; \bR^\star \sim \text{N}(\bzero, Q_{S; \tau_2^\star, \lambda^\star}^{-1}),
	$ 
	and we fix $\tau^\star = 4.5$, $\tau_2^\star = 2$, $\lambda^\star = 500$, and $\zeta_i^\star = 0.5\exp\{i^{0.05}\} - 1.5$. 
	For {\textbf{Simulation 2}}, we consider a similar scenario to Simulation 1 but with a much smaller proportion (5\%) of the locations having strong predictive power, whereas the rest of the locations do not have any predictive power. The \texttt{GPDA} model is again correctly specified. We fix $\tau^\star = 1.5$, $\tau_2^\star = 2$, $\lambda^\star = 500$, and $\zeta_i^\star = 0.5\exp\{i^{0.05}\} - 1.5$. 
	For {\textbf{Simulation 3}}, we assess the performance of the methods when the locations are mutually independent, the noise variances are equal between groups, and a small proportion (10\%) of the locations are weak signals, i.e. $\texttt{VLDA}$ is correctly specified. This is a boundary case whereby the true log length scale $\bnu_i \rightarrow - \infty$.  Lastly, {\textbf{Simulation 4}} allows us to assess the performance of the methods when the $\text{GPDA}$ model is misspecified. In particular, the true covariance matrix has a uniform structure with all diagonal entries equal $1$ and the off-diagonal entries equal $0.95$. A small proportion (10\%) of the locations have strong predictive power.

	\begin{figure}[!t]
		\centering
		{\includegraphics[width=0.8\textwidth]{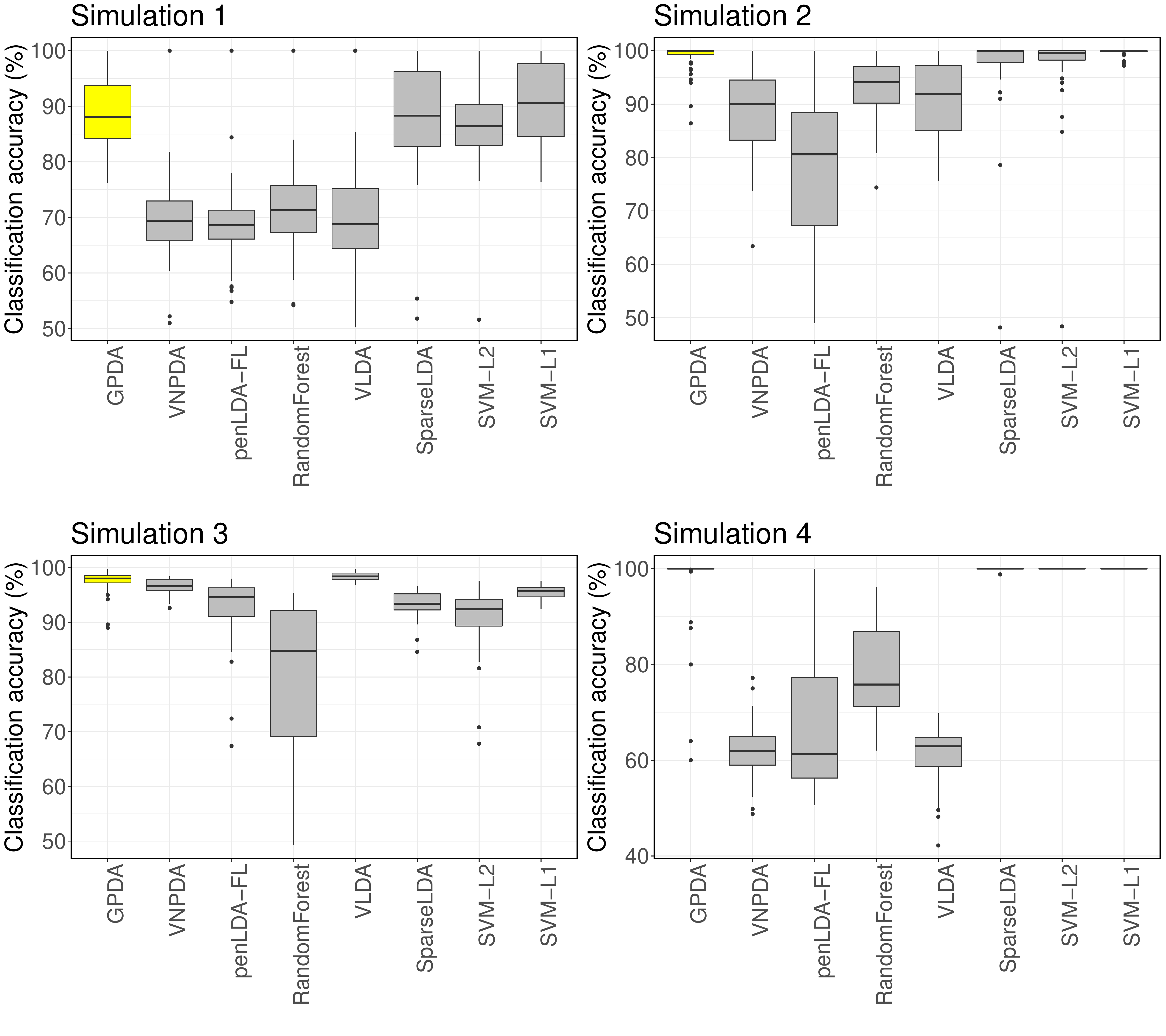}}
		\caption{Classification errors rates (\%) for Simulations 1 to 4.}
		\label{fig:SimResultClassErr}
	\end{figure}
	
	\begin{figure}[!t]
		\centering
		{\includegraphics[width=0.8\textwidth]{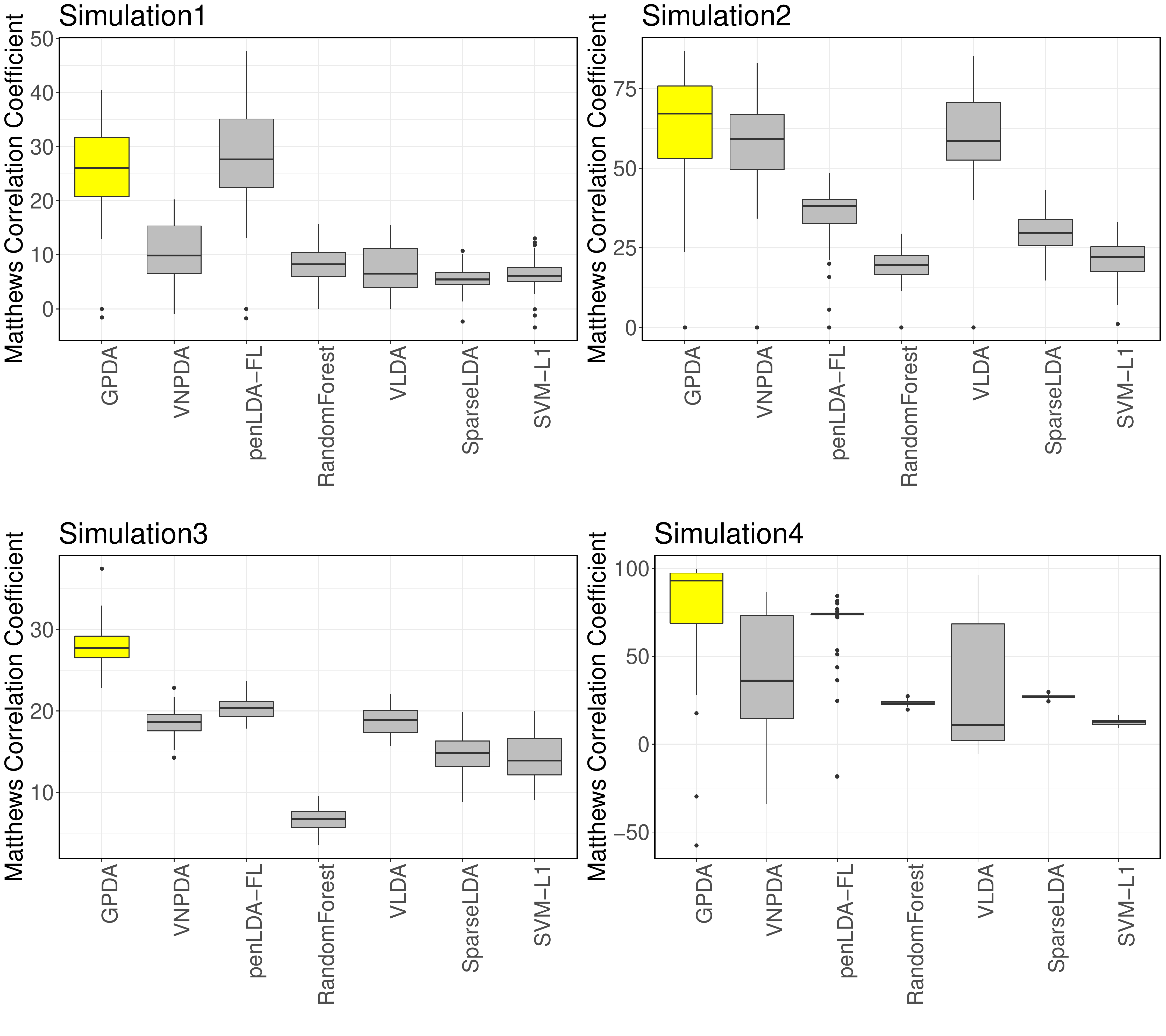}}
		\caption{$MCC \times 100\%$ for Simulations 1 to 4.}
		\label{fig:SimResultMCC}
	\end{figure}
	\paragraph{Results.}
	We assess the classification performance of the methods using the classification error rate, true positive rate and true negative rate, and the variable selection performance using the Matthews correlation coefficient. Results are provided in Figures \ref{fig:SimResultClassErr} and \ref{fig:SimResultMCC}, with further details in the Supplementary Material.  \texttt{GPDA} achieved good classification and variable selection in comparison to the alternative methods. In Simulation 1, \texttt{GPDA}, \texttt{SVM-L1}, and \texttt{SparseLDA} attained comparably high classification accuracies. Moreover, \texttt{GPDA} attained the second highest mean MCC, whereas \texttt{SVM-L1} and \texttt{SparseLDA} did not perform well for variable selection. \texttt{penLDA-FL} outperformed \texttt{GPDA} in feature selection in this simulation setting as it is well-known to perform well as a feature selector when the signal strengths are weak. In Simulation 2, \texttt{GPDA} and \texttt{SVM-L1} attained comparably low classification errors, while \texttt{GPDA} yielded the highest mean MCC. In Simulation 3, \texttt{GPDA} achieved the second lowest mean classification error rate. This demonstrates its ability to perform well even in the case when a simpler model fits the data well. Moreover, it outperforms \texttt{VLDA} in feature selection as the Ising prior works well when the true discriminative process $\bgamma^\star$ is smooth. In Simulation 4, \texttt{GPDA}, \texttt{SVM-L1}, \texttt{SVM-L2}, and \texttt{SparseLDA} achieved comparably high classification accuracies, while \texttt{GPDA} achieved the highest mean MCC. This demonstrates our proposed method's robustness when the true precision matrix is not sparse.
	
	\subsection{Proteomics datasets}
	
	We consider two datasets that aim to predict and identify markers of 1) SARS-CoV-2 in nasal swabs using matrix-assisted laser desorption/ionization (MALDI) MS \citep{nachtigall2020detection} 2) breast cancer in plasma using surface-enhanced laser desorption and ionization (SELDI) protein MS \citep{shi2006declining}. We assess the performance via five-fold cross validation (CV) classification accuracies. Moreover, for methods that perform variable selection, we compute the variable selection rate at each location. To investigate the utility of a high-dimensional approach versus the traditional approach in bioinformatics, we include a comparison of the competing methods with \texttt{LDA-traditional} and \texttt{QDA-traditional} as benchmark methods. These methods are similar to the traditional approach in that they first employ a peak detection method to identify peak locations, and followed by fitting a low-dimensional classification model using the identified peak locations.

	
	\paragraph{Data description.} 1) COVID-19 has shaken up the world; in only a year and half from its first appearance, approximately 190 million people have been infected, over four million have died, and over half of the world population has experienced some form of lockdown\footnote{John Hopkins Coronavirus Resource Center \url{https://coronavirus.jhu.edu/}}. To improve testing capacity in countries that lack resources to handle large-scale PCR testing, the SARS-CoV-2 dataset was collected using equipment and expertise commonly found in clinical laboratories in developing countries. The dataset contains samples from 362 individuals, of which 211 were SARS-CoV-2 positive and 151 were negative by PCR testing. 
	The processed spectra contain $T= 25,001$ variables and are depicted in Figure \ref{fig:covid_spectra}. 2) Breast cancer is a common and deadly disease, and improvements in early detection and screening are needed for improved treatment and survival. Towards this goal, this dataset was collected to investigate and identify markers from plasma that discriminate between controls and breast cancer patients. The processed spectra contain $T=10,451$ variables. Due to heterogeneity in breast cancers, in the following, we focus on discriminating between healthy controls and HER2 (with $n=119$). 
	More details on the raw data and required processing steps for both examples are in the Supplementary Material.
	\begin{figure}[!t]
		\centering
		{\includegraphics[width=1.0\textwidth]{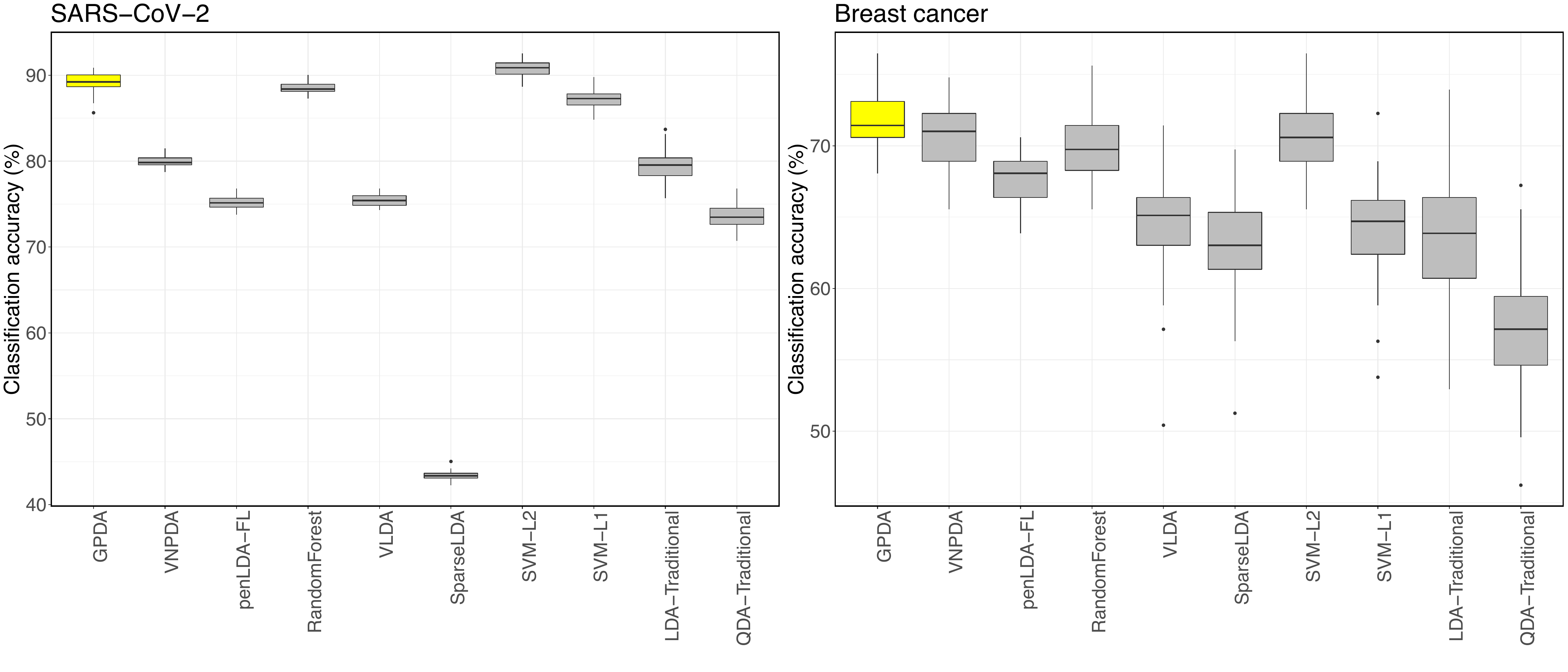}}
		\caption{Classification accuracy (\%) for SARS-CoV-2 (left) and breast cancer (right).}
		\label{fig:CombineClassAcc}
	\end{figure}
	\begin{figure}[!t]
		\centering
		{\includegraphics[width=1.0\textwidth]{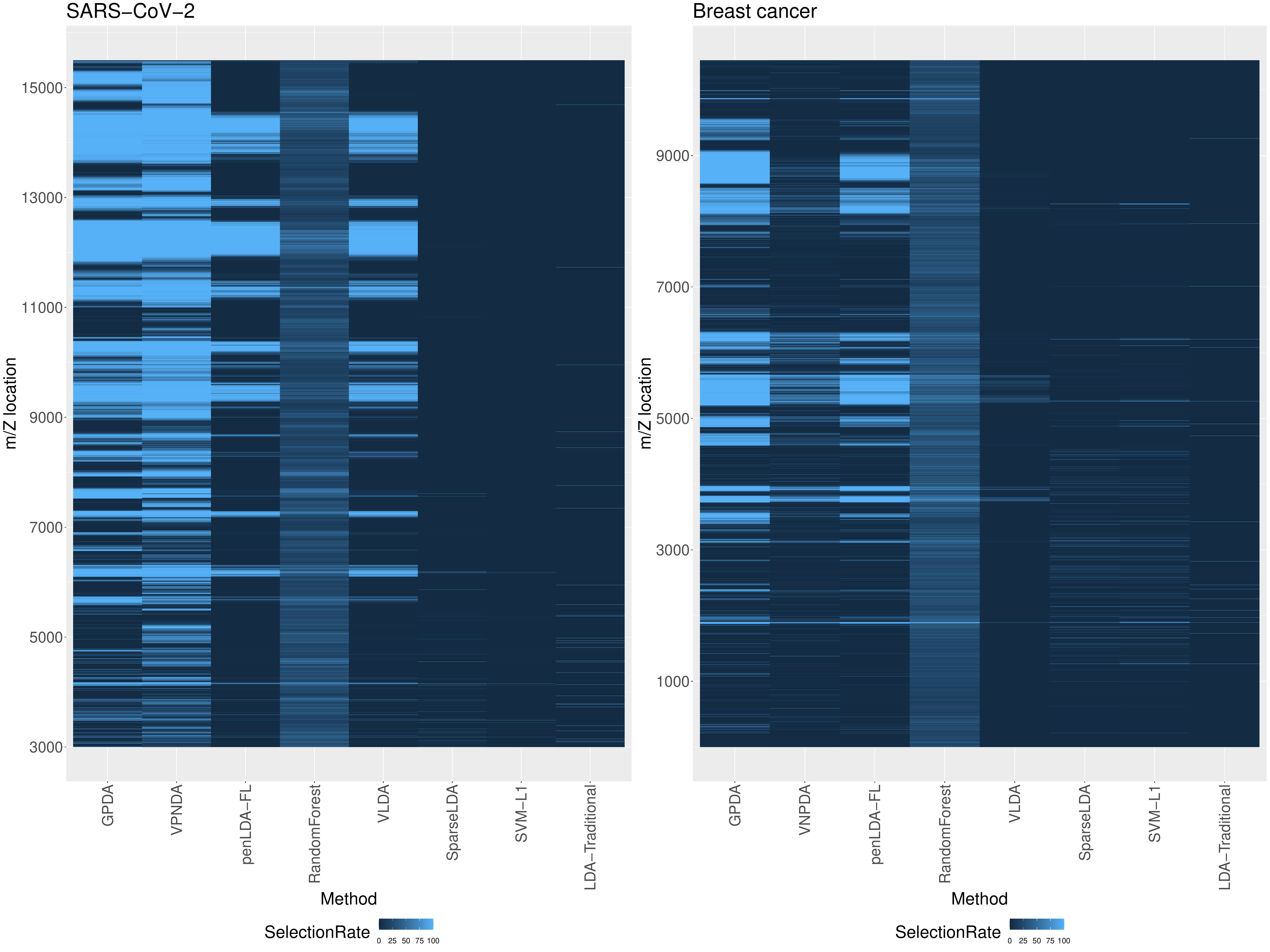}}
		\caption{Variable selection rate for SARS-CoV-2 (left) and breast cancer (right). }
		\label{fig:CombineVarbSelect}
	\end{figure}
	
	\paragraph{Results.} Classification accuracies of all methods for breast cancer and SARS-CoV-2 are presented in Figure \ref{fig:CombineClassAcc}. Other supporting plots can be found in the Supplementary Material. In both datasets, \texttt{GPDA} attained amongst the highest classification accuracy, with high true positive and true negative rates (refer to the Supplementary Material). \texttt{GPDA}'s good classification performance can be attributed to its ability to account for the highly non-stationary correlation structure that is evident from the location-varying roughness in the observed spectra in Figures  \ref{fig:covid_spectra} and S6 (Supplementary Material). Moreover, the  posterior expected log length-scales (Figures S8 and S10 in the Supplementary Material) exhibit an overall increasing trend that is congruent with the spectra being flatter at higher values of $m/z$. We also observe that methods which unify variable selection and classification in a single framework generally performed better than the traditional peak detection methods, suggesting that the two-stage peak-detection approach leads to a substantial loss of information for both datasets. Figure \ref{fig:CombineVarbSelect} summarizes the variable selection frequencies. Note that although both \texttt{GPDA} and \texttt{SVM-L1} performed well in terms of classification accuracy in the SARS-CoV-2 dataset, \texttt{GPDA} has identified many more locations as discriminative locations than \texttt{SVM-L1}. This disparity may be attributed to a subtle difference in the variable selection component of both methods. Specifically, the variable selection component in \texttt{GPDA} identifies all discriminative locations, whereas the $L_1$ penalization in \texttt{SVM-L1} identifies the optimal set of locations that minimizes classification error. We see a different trend for the breast cancer data where the classification accuracy of \texttt{SVM-L1} is clearly lower. For both datasets, \texttt{SVM-L2} attains mean accuracy comparable to \texttt{GPDA}.  However, unlike \texttt{SVM-L2}, \texttt{GPDA} identifies discriminative locations and hence may be used to identify differences in protein compositions. This information may be useful for developing new diagnostic tools and an effective anti-viral treatment. The other competing methods considered showed less comparable performance. 

	\section{Conclusion}
	\label{sec:conc}
	In this paper, we developed a comprehensive and unified framework for classification and variable selection with high-dimensional functional data.
	To account for the non-stationary and rough nature of the realized functions, we introduced a two-level non-stationary GP model with carefully chosen kernel structures, and in combination with the DA framework. Moreover, this is coupled with an Ising prior to allow for smoothness in the variable selection component. The model poses serious computational challenges due to its complexity and high dependence between parameters, hierarchical layers, and latent variables. To deal with these challenges, we proposed an inference scheme that exploits a number of advances in GPs and variational methods, as well as computational tricks, to improve the scalability and 
	efficiency of our entire inference algorithm from $O(nT^3)$ to $O(nT)$. The performance of our approach in comparison to competing methods is demonstrated in several simulated and real MS datasets. Results indicate that our method performs consistently as the best or second best in all scenarios. In addition, for the proteomics data, we illustrated how combining the steps of peak detection, feature extraction, and classification into a unified modeling framework, that accounts for the complicated dependence structure in both the inputs and variable selection, outperforms the traditional two-stage approaches commonly used in practice. We focused in this work on one-dimensional functional data, however future work will explore extensions for multivariate
	structured functional inputs, e.g. images. Other choices of kernels and deeper architectures can also be implemented to account for smoother and/or more complex structures arising in other applications.
	
	\section*{Supplementary material}
	
	All supplementary content and codes for this article may be downloaded from the repository: \url{https://github.com/weichangyu10/GPDAPublic}.

	
	\bibliographystyle{apalike}
	\bibliography{references.bib}
\end{document}